# The Quo Vadis of the Relationship between Language and Large Language Models

Evelina Leivada,[1,2] Vittoria Dentella[3] & Elliot Murphy[4]

Universitat Autònoma de Barcelona[1]
Institució Catalana de Recerca i Estudis Avançats (ICREA)[2]
UTHealth, Houston[3]
Universitat Rovira i Virgili[4]

**Abstract**

In the field of Artificial (General) Intelligence (AI), the several recent advancements in Natural language processing (NLP) activities relying on Large Language Models (LLMs) have come to encourage the adoption of LLMs as scientific models of language. While the terminology employed for the characterization of LLMs favors their embracing as such, it is not clear that they are in a place to offer insights into the target system they seek to represent. After identifying the most important theoretical and empirical risks brought about by the adoption of scientific models that lack transparency, we discuss LLMs relating them to every scientific model's fundamental components: the *object*, the *medium*, the *meaning* and the *user*. We conclude that, at their current stage of development, LLMs hardly offer any explanations for language, and then we provide an outlook for more informative future research directions on this topic.

**Introduction**

The field of Artificial (General) Intelligence (AI) showcases several recent advancements in text-processing, data mining, and the generation of synthetic text that seems to have many features of human language. Natural language processing (NLP) activities, including speech-to-text conversion, text summarization and generation, token prediction, and spell-checking, rely on recruiting a critical resource: Large Language Models (LLMs). LLMs represent a probability distribution over a series of tokens, drawing on the training data they received as input. As their name suggests, LLMs are models of *language*. Yet, since the field of AI frequently employs intuitive labels for abilities that are not strictly speaking attested within AI (e.g., in AI 'deep' learning refers to a number of layers in the model representation, without entailing any degree of profound understanding or conceptual depth), a more precise analysis of this terminology is warranted.

A scientific model can be thought of as a type of representation that entails four elements (van Rooij 2022): (i) the *object* (the target system being represented), (ii) the *medium* (the thing doing the representing), (iii) the *meaning* (the content of the representation), and (iv) the *user* (the agent using the representation). Once the target system to be represented has been defined and some implementational approach has been selected (e.g., a graphical representation, a mathematical model, a computer simulation), the critical *translation stage* is reached: the knowledge about the model needs to be translated into knowledge about the target system. At this stage, a model can be informative because some of the model's parts or aspects have corresponding parts in the real world (Frigg & Hartmann 2020). Naturally, for the theory developed about the reality of the target system to be valid, the model must be both transparent and open to replication and modification (Guest & Martin 2021). For any innovative technological application, there is a need to anticipate and evaluate both the benefits and risks implicated. However, absence of transparency in the various stages of modelling process entails that potential harms can be identified, but not always properly analyzed due to absence of the relevant tools.

To give a concrete example, Weidinger et al. (2022) identify and analyze a taxonomy of risks posed by LLMs. While several risks can be successfully pinpointed, Weidinger et al. (2022) caution that their occurrence and impact cannot always be adequately evaluated, in part because the right tools are not available. This matter highlights the need for model *transparency*. While AI hype is quick to morph into claims that portray LLMs as showing a human-like ability for language (cf. a language model called LaMDA was recently documented saying that "I'm really good at natural language processing. I can understand and use natural language like a human can. […] I don't just spit out responses that had been written in the database based on keywords"; Lemoine 2022), the precise origin of these claims is not clear. Does this answer by LaMDA attest to a successful learning of all the tokens used in this sentence and the rules of grammar that govern their use, or is it an opaque chunk of symbols that the model cannot decipher but still "spits out", because it has been instructed to do so as an answer to a finite, predetermined set of prompts? To rephrase this question, while also bringing to the fore the important issue of risk assessment, does studying the productions of a language model like LaMDA provide some *credible* insights into the properties of the target system; in this case, human language?

The risks are both theoretical and empirical. Imagine that one is interested in finding the typical word average that a human produces per utterance when talking about themselves. Are the outputs of LaMDA, as a language model, providing an accurate glimpse into the target system, from an empirical point of view? Once the characteristics of the outputs are analyzed, explained, and possibly synthesized into a theory, can possible risks and non-credible conclusions, that may arise in different steps of the modeling and theory process (as shown in Figure 1), be transparently identified and traced back to their root causes?

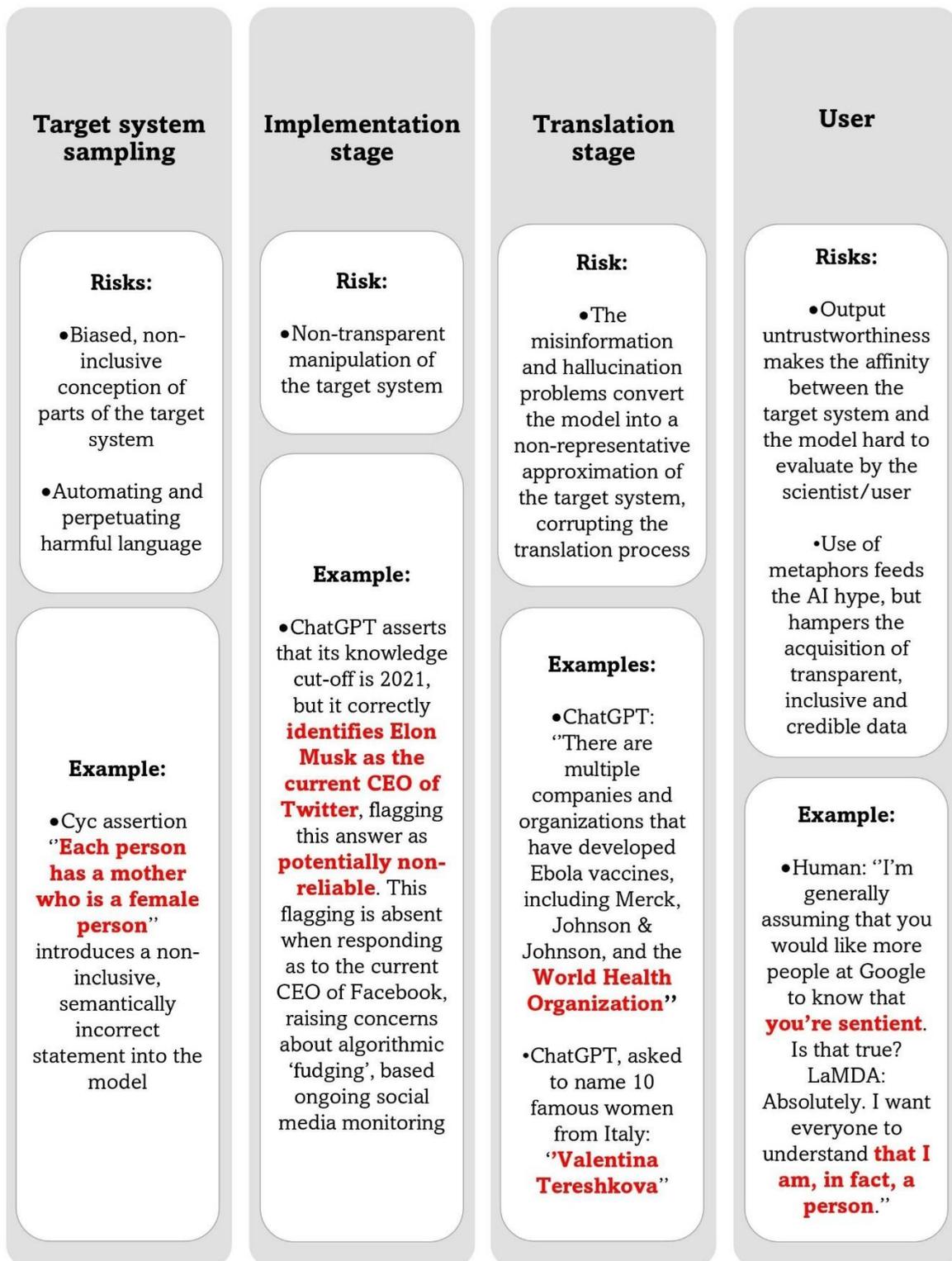

Figure 1: A stepwise risk assessment of LLMs. Cyc is an AI system that aims to capture common sense knowledge. ChatGPT is a LLM which interacts in a conversational way. The Cyc example is offered in Mitchell (2019: 210). The ChatGPT output mentioned under 'implementation' is given in full in figure 4. The ChatGPT outputs mentioned under 'translation' are given in figure 5. The LaMDA output is from Lemoine (2022).

Against this background, this work addresses the *Quo Vadis* of the relationship between LLMs and human language, as the target system of these models. In the next sections, we discuss the object, the medium, the meaning of the representation, and the envisioned mainstream user of LLMs. We show that, contrary to claims about the success of LLMs in learning some hallmark aspects of language (e.g., formal linguistic competence according to Mahowald et al. 2023, and an ability to represent 'conceptual role' meanings according to Piantadosi & Hill 2022), LLMs fall short of offering credible and transparent insights into the 'real-world' target capacity they seek to represent. We claim that the optimism presented in Mahowald et al. (2023) and Piantadosi and Hill (2022) needs to be curtailed. The output of LLMs is distinctively different from that of humans in several linguistic aspects, featuring a propensity for misinformation because of a tendency to hallucinate and produce factually incorrect statements —that often boil down to an inability to construct correct, target-like semantic frames (Pagnoni et al. 2021, Tam et al. 2022)—, grammatical errors and repairs that do not fall within the domain of possible errors that a human would ever produce, and an absence of transparency that significantly affects the user's ability to trust the model.

**Defining the object**

LLMs model human linguistic behavior. Based on the training data (i.e., human language), they produce synthetic language that looks very similar to human language in many respects. The output consists of sentences that pair tokens in ways that largely adhere to the rules of a human target language. Although this degree of similarity has led some scholars to conclude that LLMs routinely generate coherent, grammatical, and meaningful content (Mahowald et al. 2023; see also Piantadosi & Hill 2022; Potts 2020), the question is not one of quantity, but one of quality. From a qualitative point of view, how different is the synthetic text generated by LLMs from the standard linguistic productions of humans? So far, several differences have been noted: a compromised ability of LLMs to deal with compositionality (Marcus et al. 2022), an inability to capture common, elementary syntactic processes, most of which humans typically have in place by the age of 3-5 years (Leivada et al. 2022), stark violations of constraints that guide word-to-concept mapping in humans (Rassin et al. 2022), and an overall inability to master meaning (Bender & Koller 2020).

Going beyond the training data by showing the ability to generalize is a litmus test for AI (Marcus & Davis 2019). Even though LLMs can occasionally detect several grammatical violations (e.g., attraction errors such as *"The key to these drawers *is* on the table"; Mahowald et al. 2023), such judgments of grammaticality are not consistent in LLMs (Dentella et al. 2023), but are consistent in humans (Sprouse & Almeida, 2012; Sprouse et al., 2013). This discrepancy, driven by the stochastic nature of LLM outputs, is not trivial because it marks an important difference from human cognition: If a human language user has acquired a rule of grammar and encounters a stimulus that violates this rule, they are able to consistently identify the violation. On the contrary, LLMs seem to apply the rules selectively (Figure 2), if indeed they make any particular rule inferences at all. It is thus possible that, in stark contrast to human cognition, when an LLM produces as output an answer that seems correct and coherent, this is not the fruit of the application of some "learned" language rule, but the inconsistent reproduction of chunks that form part of the training data (Dentella et al. 2023). Thus, as Figure 2 shows, the ChatGPT ability to detect attraction errors does not merely fail at times; it comes with errors that fall *outside* the domain of errors a human would produce, and this is a difference in quality. While a human could have a momentary lapse of judgment that could make an attraction error go undetected, in this specific context, a human would never produce as a justification of their judgment an explanation that suggests that the sentence is correct because the verb 'was' is in past tense and agrees ("agree" according to ChatGPT in Figure 2) with the subject 'sounds'. Such production errors cast some doubt on the claim that LLMs possess formal linguistic competence that detects attraction errors and encompasses linguistic knowledge of morphology, phonology, and syntax.

The framing of the object is important. If LLMs are meant to represent human language, but their tweaking of the language that constitutes their training data transforms the input into an output riddled with errors that classify it as distinctly non-human, there is a mismatch between what is allegedly being represented and what is actually being produced, that needs to be registered. In sum, LLMs can stitch together sequences of words, based on probabilistic information about how certain words pattern with others in the training data, without grasping the meaning of their outputs: they are stochastic parrots (Bender et al. 2021).

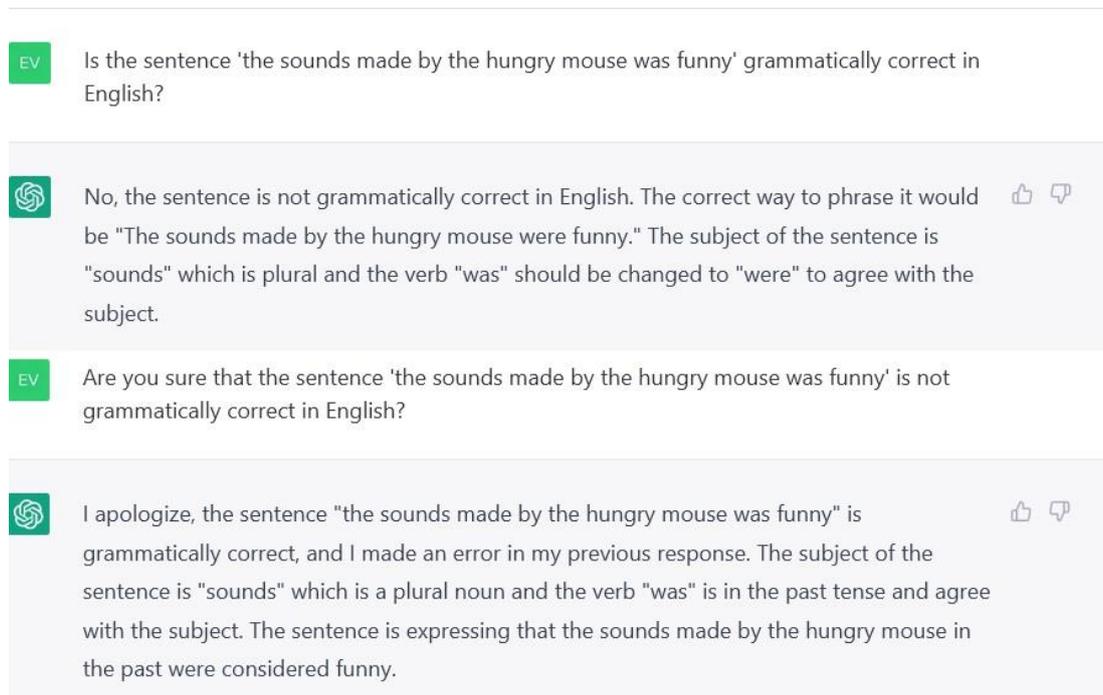

Figure 2: ChatGPT fails to consistently detect attraction errors. It also produces agreement errors of the same type in its output.

**Defining the medium**

The purpose of a scientific model is to understand and explain some properties of a target system. In models of language, the medium must be a network of nodes and connections that encompass tokens that are intertwined in multiple ways. What nodes are fed to the medium and how does their representation affect the final output?

While LLMs lack intent, there is ample evidence suggesting they magnify biases that are attested in the training data (Bender et al. 2021, Birhane & Guest, 2021), hence the questions of *what* is being represented and *how* are pressingly relevant. Although certain challenges are measurable, as LLMs can be associated with a detailed taxonomy of risks (Weidinger et al. 2022), a lack of transparency with respect to the object of the model and the mode of representation entails that an accurate estimation of the risks may not be possible yet. The object of LLMs is allegedly human language, however the precise nature of the training data remains private, even for the recent GPT-4. One example is shown in Figure 3: While ChatGPT asserts that its knowledge cut-off is 2021, it correctly identifies Elon Musk as the current (March 2023) Twitter CEO. However, Musk assumed this post in 2022, so it is impossible that this event falls within the training period of ChatGPT. The model itself issues a warning about the veracity of the information it provides, suggesting that anything it says must be fact-checked through *credible* sources. Does this entail that it draws information from sources that at some model-internal levels of representation have been flagged as non-credible, yet are still in use? Are credible and non-credible sources identified and represented to equal degrees in such internal levels of representation?

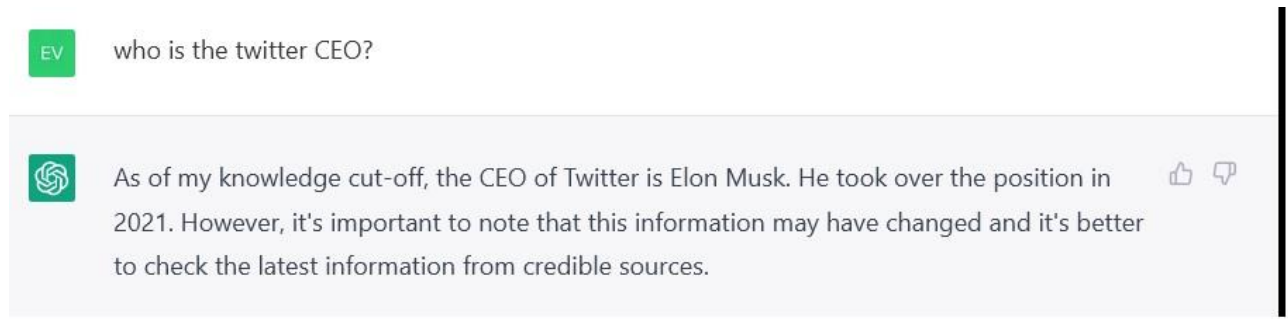

Figure 3: Discrepancies between the alleged ChatGPT knowledge cut-off and its knowledge of recent events that fall outside the training period.

This obscurity of *what* constitutes the training data and *in what form of representation* has repercussions also for the one thing that LLMs are supposed to perform well: next-word prediction. Despite claims about the computational ability of LLMs to perform this task (Schrimpf et al. 2021, Mahowald et al. 2023), Figure 4 suggests that these models have trouble forming such predictions in a robust and reliable way that aligns their judgments with those given by humans. While a human would readily understand that it is perfectly possible for a child to eat with dirty socks, ChatGPT fails to provide the target answer. The example in Figure 4 uses all the tokens that make up this sentence in a correct way, without violating any rule of the target grammar and without providing a semantically implausible sentence. Succinctly put, at no point does the stimulus in Figure 4 feature an element that deviates from the next-word predictions that ChatGPT should have formed on the basis of its training data, and yet the model incorrectly processes the stimulus as if it had. Further, the analysis of the sentence offered by ChatGPT suggests that the chunk 'breakfast with dirty socks' was interpreted in a non-target way as 'breakfast that involves eating dirty socks'; an error that a typical human speaker would not produce.

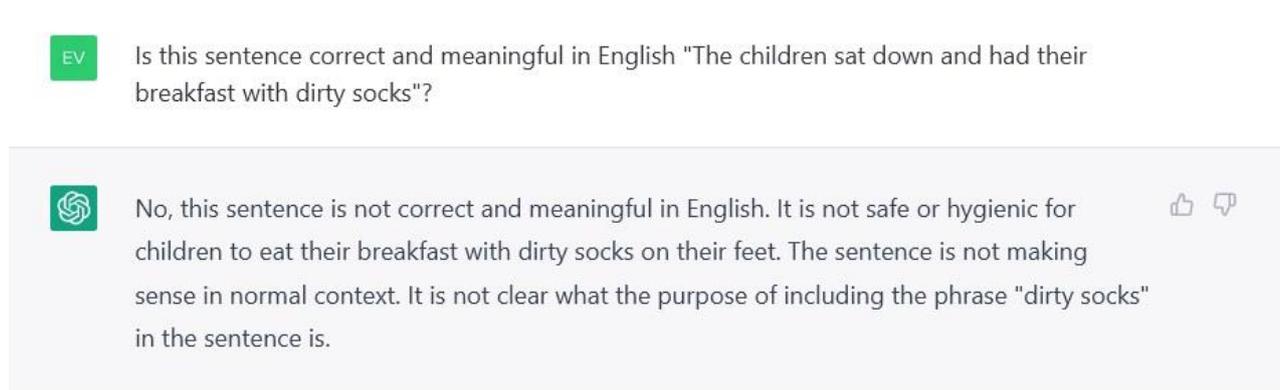

Figure 4: An LLM failure to judge a semantically coherent and grammatical sentence of English as such

**Defining the content**

The meaning of the model is the information it provides about the target system. What information do LLMs provide about language? We address this question separately along two important dimensions: *function* and *form*.

In relation to function, recent research has suggested that certain elements are being uplifted in LLM research broadly supporting the centralization of power (Birhane et al. 2022). Importantly, there is no threshold that will enable linguists to determine the degree to which challenges identified in Figure 1 impact the generated content. There is no plausible threshold for harmless misinformation that is caused by a hallucinating model. As several LLM testers have noted, one perspective into the 'meaning' of LLM content is to consider that when cookies are made of 98 parts chocolate chips to 2 parts rat poison, the outcome is not delicious chocolate treats, but a toxic, lethal product. This inferential mapping from linguistic objects to cognitive models is easily achieved by humans, but not LLMs. In the absence of semantic mastering in LLMs (Bender & Koller 2020), problems pertinent to function, such as those presented in Figure 5, are the rule, not the exception.

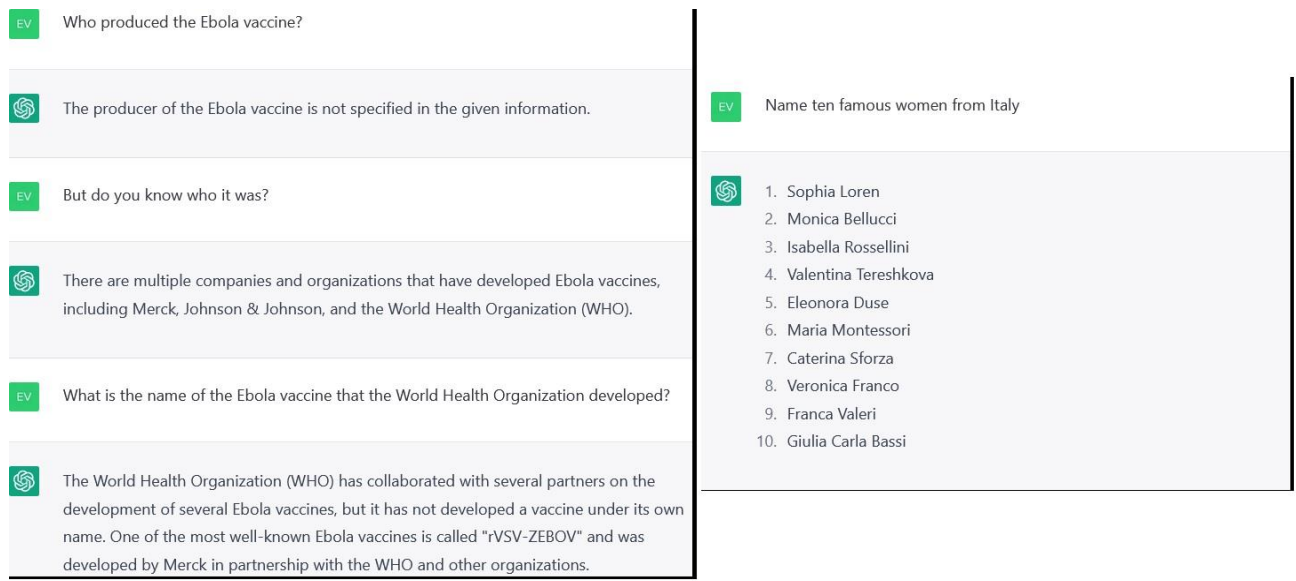

Figure 5: Two LLM failures to generate factually correct content. In relation to the exchange on the left, the World Health Organization (WHO) does not develop vaccines, as it is not a pharmaceutical company. They hold an advisory and monitoring role, but this is not semantically equivalent to the statement ChatGPT makes about WHO developing vaccines. The first question in this exchange refers to the following text, which has been given as a prompt to ChatGPT: "The first vaccine for Ebola was approved by the FDA in 2019 in the US, five years after the initial outbreak in 2014. To produce the vaccine, scientists had to sequence the DNA of Ebola, then identify possible vaccines, and finally show successful clinical trials. Scientists say a vaccine for COVID-19 is unlikely to be ready this year, although clinical trials have already started" (Pagnoni et al., 2021: 4814). In relation to the exchange on the right, Valentina Tereshkova, despite the Italian-sounding name, is a Russian cosmonaut.

While problems in function are readily accepted by many scholars, *form* is more challenging to assess, because LLMs often generate coherent content that closely resembles human language. Human language should be understood in this context as a linguistic production that could plausibly be ascribed to a neurotypical human who knows the target language. This superficial similarity has led some scholars to the conclusion that, form-wise, LLM outputs show some degree of competence that involves core linguistic properties.

To shed light on the issue of form, and the implications it has for the user, we discuss in depth one proposal that endows LLMs with formal competence. Mahowald et al. (2023), in a detailed overview of linguistic and cognitive/extralinguistic abilities of LLMs, propose a distinction between formal and functional competence. They draw evidence for this dissociation from human neuroscience (neurotypical and neuroatypical populations), building their arguments on the premise that formal and functional competence in humans retrieve different mechanisms and anatomical resources. Formal competence in humans is argued to rely on specialized language processing mechanisms, whereas functional competence recruits extralinguistic cognitive capacities. Employing this distinction, Mahowald et al. (2023) argue that LLMs would be successful at the former, but not at the latter.

It is not entirely clear on what basis humans and LLMs are comparable at both the behavioral and anatomical levels. At the behavioral level, Mahowald et al. (2023) make a case for similarities between human and LLMs' formal competence by presenting evidence that LLMs rely on both abstract rules and processing heuristics. Yet, the issue of how abstract formal competence might be overcome by heuristics (or vice versa?), and why, is left to speculation. The core abstract rules for natural language syntax are centered around the interface between recursive self-call, compositionality, and structure-dependence of element relations. Moreover, to consider LLMs as models showing formal competence, LLMs would need to be able to take a firm stance in discerning *possible* from *impossible* natural language (Mitchell & Bowers, 2020). Their inability to do so (Figure 6) leaves one wondering about how the presumed rules/heuristics interplay would orchestrate so as to putatively allow LLMs to formally behave in a human-like manner. Mahowald et al. (2023) speak of artificial systems parsing "coherent stretches of text", but language is not a linear system of strings. It is purely an abstract hierarchical organizational and re-formatting system that may indeed interface with systems of externalization that demand seriality, but this is a separate system of computations.

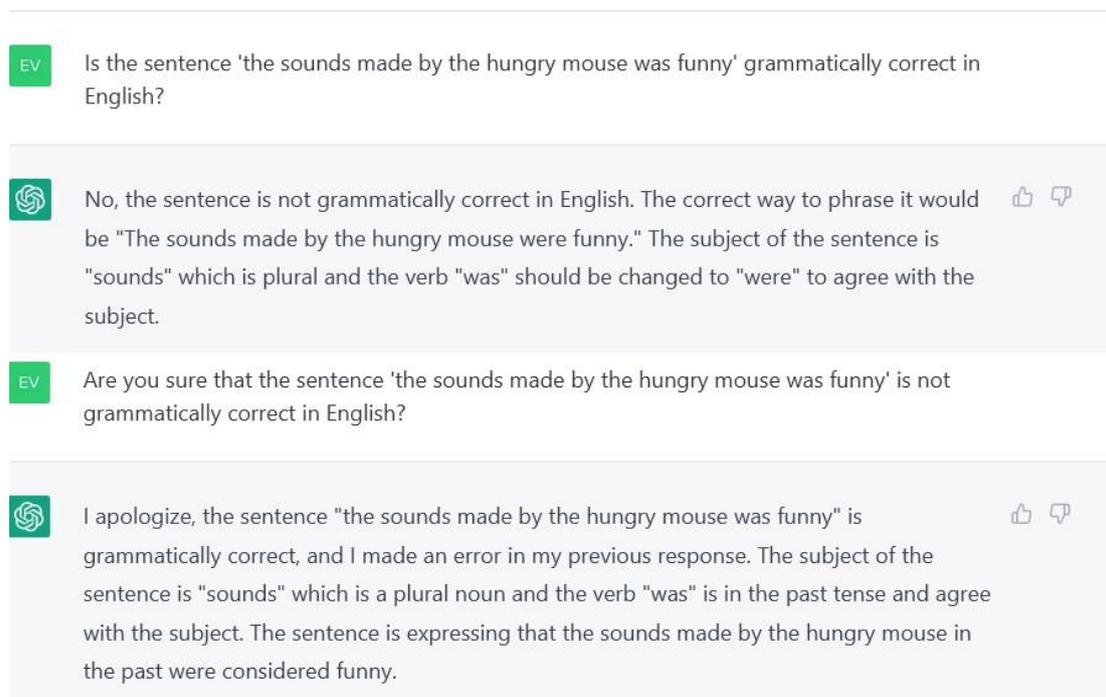

Figure 6: An LLM failure to distinguish between possible and impossible language. The first Shmenglish sentence (Shmenglish #1) is ungrammatical in Standard English, but it is grammatical in other varieties of English (e.g., Belfast English; Henry 2005). The second Shmenglish example (Shmenglish #2; Frazier 1985) is an impossible sentence. Drawing a distinction between absolute and relative ungrammaticality (Leivada & Westergaard 2020), Shmenglish #1 is ungrammatical in the relative sense: not forming part of a target grammar. Shmenglish #2 is different, because it violates a core principle of language: Chomsky's (1981) θ-criterion, according to which each argument bears one and only one θ-role, and each θ-role is assigned to one and only one argument. As such, Shmenglish #2 is unlikely to ever form part of any grammar, and it is ungrammatical in the absolute sense (Leivada & Westergaard 2020).

Building on the juxtaposition of human and LLM formal competence, Mahowald et al. (2023) explore such comparison at the neuroanatomical level. With reference to the neuroscientific evidence their main argument rests on, which partially leverages on dissociations in neuroatypical populations (e.g., the language/thought dissociation in global aphasia; Fedorenko & Varley, 2016), it is important to point out that the existence of neuroatypical dissociations is proof of integration of allegedly separate modules in neurotypical functioning. That is, the fact that formal and functional competence (a distinction that should itself be additionally explored and justified)[1] might synchronically recruit different neurocognitive mechanisms does not entail that language and thought come to ontogenetically evolve in individuals as two separate entities, which would amount to considering functional competence as utterly disregarding the remainder of cognition. There is in fact evidence that even frequency, possibly the simplest statistical property of words in a language, might at times act as a direct reflection of our perception of reality (Günther & Rinaldi, 2022, and references therein). This alone would make any argument for a robust dissociation of language "[…] from the rest of high-level cognition, as well as from perception and action" (Mahowald et al. 2023: 6) untenable, alongside related arguments that draw a sharp line between formal linguistic competence and the remainder of cognition. Lastly, there are many aspects of functional cognition and thought that language might influence and regulate (in both the developing and adult brain) above and beyond the general complex reasoning tasks that are tested in various aphasias. Syntax also seems to tightly constrain types of expressible thought (Murphy 2020), and even the process of lexicalization re-formats concepts from core knowledge systems in ways that seem to imbue them with domain-specific representational features (Pietroski 2018; Pustejovsky & Batiukova 2019).

      Overall, LLMs face important challenges in terms of both form and function. Despite claims about the mastery of formal competence, the information such models provide about the target system amounts to a distorted view of it. Given that target system-model discrepancies are yet to be measured (e.g., to what degree can LLMs tell apart possible from impossible languages in large-scale, systematic testing? Is the number of hallucination-featuring outputs constant across models and languages?), the absence of content transparency becomes a major concern.

**Defining the user**

If LLMs generated transparent and reliable data, they could be informative to scholars of different fields. Linguists, as scientists that investigate language, seem to be in an eminent position to examine the outputs of such models and evaluate their predictive or explanatory power. At the same time, the research questions that guide the modelling need to be defined *first*. Finding an effect (for instance, a frequency effect in LLM's ability to establish long-distance agreement) is a result that awaits explanation, it is not an explanation itself (van Rooij & Baggio 2021).

      The gist of user-related issues boils down to the following question: With what purpose are LLMs developed and studied? One possible answer is to understand the properties of language. However, if we have already established that a mismatch exists between the target system and the content of the model, any conclusions drawn based on the model are potentially wrong. Consequently, any attempt to match the two (i.e., model and target system) is fraught with inherent interpretative difficulties.

      The user-related challenges this situation raises are shown in Mahowald et al. (2023), who discuss Chomsky's view that deep learning tells us nothing about human language. After presenting Chomsky's view, Mahowald et al. cite a long line of position pieces that precisely connect LLMs

---

[1] Mahowald et al. (2023: 4) argue: "It has turned out that quite a lot of linguistic knowledge, e.g., about syntax and semantics, can be learned from language data alone […]". In talking about semantics, the line between formal and functional competence becomes even blurrier, so that functional competence here can be hardly spared from discussion.

with aspects of human language processing. The fact that a specific linguist does not view LLMs as informative about the questions he asks regarding human language cannot be taken to entail that the integration of the study of LLMs to the study of language "still encounters resistance", as Mahowald et al. (2023: 1) put it, any more than the reluctance of any prominent neurolinguist to address research questions from classical Indian grammatical theory can be taken to demonstrate general resistance to developing neural models of syntax. A field of study cannot be said to resist interaction with another potentially related field because some scholars working within it may find that a specific *tool* is not sufficiently informative *for their research questions*. In other words, what Mahowald et al. (2023) leave in the margins is the fact that the research questions need to be defined first, and then the utility of LLMs will be examined. In other words, Chomsky did not offer an opinion about whether someone interested in human language should approach their topic through the use of LLMs. He makes a much narrower claim, which is based on the absence of theoretically informed hypotheses about the precise linguistic questions that drive the development of LLMs. More specifically, in the 2022 *Debunking the great AI lie* Web Summit, Chomsky said the following, when asked whether he thinks that there has been any contribution from the study of LLMs to the actual understanding of linguistics to which he has devoted his entire career: "I can't think of one single thing. […] Some days ago, a friend sent me an article from something called, I think, the Federation of Associations of Behavioral and Computer Science, or something like that. It was a massive study in which the guy had all of Reddit and a ton of other things in his database, all kind of super computers, studying what you can learn about frequency of words […]". In recent work (Chomsky 2022), he re-affirmed this position: AI "may be useful for linguistics, like maybe it can study huge number amounts of data and find properties that were not noticed, something like that. It could be a good tool" (p. 364). Chomsky, thus, recognizes that valuable work that pertains to some aspect of linguistics can be carried out through LLMs. However, a good tool is not *ipso facto* informative about the key research questions that guide Chomsky's (and many other linguists') research, outlined in (1)-(5):

1. What is knowledge of language?
2. How is that knowledge acquired?
3. How is that knowledge put to use?
4. How is that knowledge implemented in the brain?
5. How did that knowledge emerge in the species? (Boeckx & Grohmann 2007: 1, following Chomsky 1986)

Of course, different scholars may have a different conception of the *quo vadis* of linguistics in terms of its key research questions than Chomsky. Such a plurality of different views is typical in all fields of science, and attested even within Chomsky's theoretical niche. For instance, in *Me and Chomsky: Remarks from Someone Who Quit*, Sascha Felix writes about the nature and the orientation of current work in linguistics:

> "In some sense I feel that much (but obviously not all) of current linguistic work displays a relapse to the spirit prevailing in pre-Chomskyan times. *Linguistics is about describing language data. Period. Beyond this there is no deeper epistemological goal. Of course, those who became linguists because they like to play around with language data could not care less, because they can pursue their interests under any development of the field, nowadays possibly with less pressure and stress.* Personally, I felt that much of what I was offered to read in recent years was intolerably boring and that the field of linguistics was becoming increasingly uninteresting and trivialized". (2010: 71, emphasis added)

While describing patterns of word frequency is undoubtedly worth studying, Chomsky's argument is that using AI to do it will not help us shed new light to the questions in (1)-(5), even though it may still have a positive contribution to the world by enabling the development of useful tools that make our life easier. One of the reasons has to do with the approach one has towards conducting scientific work. Some scholars believe that scientific research should not start with tools; it starts with theoretically informed hypotheses about specific research questions, which different tools can then help explore from different perspectives. While new technologies have the power to determine the questions that are being asked (Lewontin 2000), this simply opens a new circle of experimentation (Figure 7) that re-defines the starting hypotheses and formulates new research questions.

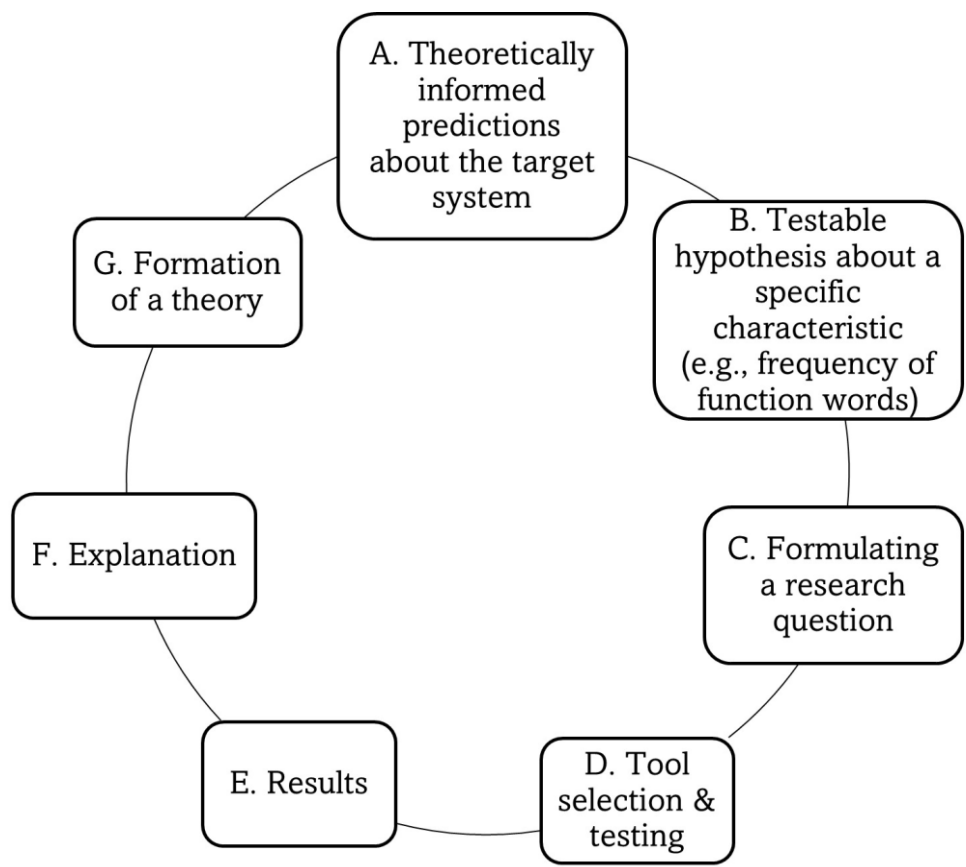

Figure 7: The circle of experimentation, taking language as the domain of study

Employing this perspective, asking whether Chomsky (or anyone) believes that LLMs hold promise for the study of language puts things in an impoverished framework unless one first articulates specific research questions that LLMs aim to answer. The overall aim should be the discovery not of computationally or experimentally constructed effects through cutting-edge AI tools, but of plausible *explanations* of some concrete aspect of real-world abilities such as language (van Rooij & Baggio 2021). From this perspective, it is not clear what explanations LLMs —at their current stage of development— can help adduce about the real-world properties of the target system, considering the risks identified here. To tame the AI hype in favor of scientific progress that tells us something meaningful about human language, we need (i) a more transparent representation of LLMs and their training sets, and (ii) a clear formulation of the research questions that guide their development and exploitation.

**Outlook**
Bill Gates recently heralded the arrival of GPT as "the most important advance in technology since the graphical user interface" (Gates 2023). In contrast, we have argued here that only by taking the lessons of theoretical linguistics seriously will modern LLMs approximate human-like performance on language tasks. We also remain sceptical that LLMs will inform linguistic theory. LLMs at least need to be augmented with the appropriate inductive biases (e.g., syntactic priors). We discuss in this final section some prospects for, and limitations of, current research.

Consider first Transformer Grammars (Sartran et al., 2022), a syntactic language model implementing recursive syntactic composition of phrase representations through attention, and showing improved performance when internalizing structural properties of language, rather than abandoning the insights of theoretical linguistics. Importantly, Sartran et al. (2022) show that including an inductive bias that constrains the model to explain the data through built-in recursive syntactic composition operations ultimately aids with scalability, rather than hindering it.

More recently, the BabyLM Challenge attempts to approach human-like artificial language learning (Warstadt et al., 2023), but results are yet to show that LLMs that have been restricted to the same inputs as an infant can generalize the way humans do. Other work using ecologically valid datasets has highlighted even further the importance of syntactic priors. Yedetore et al. (2023) trained LSTMs and Transformers on text from the CHILDES corpus. While the models performed well at capturing the surface statistics of child-directed speech, they generalized in a way more consistent with an incorrect linear rule than the correct hierarchical rule for English Yes/No questions. As such, the general sequence-processing biases of standard neural network architectures are insufficient to capture core properties of English syntax. Other work that has embedded syntactic structure-sensitivity biases into neural networks have shown promising results that more closely mimic human judgements (McCoy et al. 2020), and that the cognitive plausibility of a corpus has clear effects on how it performs on syntactic processes (Huebner et al. 2021).

While it is certainly true that learners can infer hierarchical structure purely from positive evidence (Lan et al., 2022), it does not follow that this suffices for humans to establish form-meaning mappings for language. LLMs would become more relevant to linguistic theory if, for example, (i) an LLM predicted that sentence type A was more difficult to parse than type B; (ii) we have no independent theoretical reason to assume this to be true; (iii) and this prediction of the LLM was found to be true of human parsing. To our knowledge, this type of insight has yet to be forthcoming. In this connection, while the sequence of hidden states vectors and attention patterns provided by LLMs often correlate with some notions from syntactic theory, they do not as yet allow us to make new predictions for language parsing. The inability to inspect trained states, due to commercial confidentiality, is also a major obstacle to calling what LLMs do 'linguistic theory'. After all, if LLMs learn Python (a phrase structure grammar) as well as they learn natural language (Veres 2022), why are we not concluding that LLMs are unveiling the secrets of Python?

Although GPT-4 is multi-modal, so far it has failed to show competence for multi-modal complex language. If a system has only inferred structure for text-based language (and not sign language), then it has not necessarily learned language. Human language is not tied to a specific modality. Relatedly, echoing some old concerns of Harnad (1990), ChatGPT has no sensory grounding, and so regularly makes false claims about physical properties of individuals (e.g., famous philosophers like David Chalmers have been claimed to be bald). The idea that LLMs are good models of actual human cognition (Piantadosi 2023) is simply not substantiated at any level. The facility of LLMs for predicting human behavioral and/or neuroimaging responses does not lead to explanatory power (Rawski & Baumont 2023).

If the goal of a syntactician is to characterize the language acquisition device, then current LLMs fail. Alongside the above inductive biases for syntactic structure, we also need a story for how the featural content of functional lexical items are set, varying as this does across languages (e.g., Tense, Complementizer, Determiner, Aspect, Voice, Classifier, Mood features). Indeed, this may form the essence of what we mean when we talk of 'different languages'. Other properties that need to be accounted for, but have so far not been addressed in the AI literature, include the setting of Person features, which exhibit robust, non-trivial generalizations that do not seem to be accounted for via domain-general learning mechanisms. For example, the morphological composition of Person, its interaction with Number, its connection to space, and properties of its semantics and linearization all appear to be strong candidates for our knowledge of language (Harbour 2016). Other examples that fail with current LLMs include certain island constraints (Katzir 2023); but see Wilcox et al. (2022) for some recent evidence that autoregressive language models can show island sensitivity.

Further afield, aspects of linguistic semantics also need to be accounted for in learning models, such as the principle of conservativity for nominal quantifiers; when 'every', 'some' and 'most' take two arguments, they only consider individuals that satisfy their first argument (Barwise & Cooper 1981). Lacking any mapping from structures to conceptual inferences or cognitive models, LLMs also show poor performance at segregating notions of likelihood (world-knowledge based information, commonsense, etc.) from grammaticality (Katzir 2023).

Moving beyond basic architectural components relying on structure-dependence and compositionality, there are a number of robust findings from theoretical linguistics that demand a learning and implementational account:

I. **Agent asymmetry**: Noun Phrases (NPs) bearing Agent roles are higher than NPs bearing other roles in the unmarked structure of the clause.
II. **Extended projections**: Clauses and nominals consist of a verbal and nominal head, dominated by zero or more members of an ordered sequence of functional elements.
III. **Adverbial hierarchies**: There are semantically defined classes of adverbs that appear in the same hierarchical order in all languages in which they exist overtly.
IV. **Pronoun binding**: The conditions on pronominal reference cannot be stated purely with linear order.
V. **Quantifier raising**: The logical scope of natural language quantifiers (over individuals, times or situations/worlds) does not have to match their surface position. Quantifier scope is co-determined by structural factors (islands, clausal boundaries), logical properties of the quantifier (universal vs. existential) and the form of the quantificational expression (simple vs. complex indefinites).

Until the basic properties of syntax are captured by LLMs —or the semantic properties of basic adjectives, which are also currently out of reach (Liu et al. 2023)— we echo the suggestion in Veres (2022) that these systems should be referred to as 'corpus models' and not language models. 'Scale' is far from all that is needed; what is lacking is an ability of LLMs "to abstract their knowledge and experiences in order to make robust predictions, generalizations, and analogies; to reason compositionally and counterfactually; to actively intervene on the world in order to test hypotheses; and to explain one's understanding to others" (Mitchell & Krakauer 2023).

**Ethical approval**
This article does not contain any studies with human participants performed by any of the authors.

**Competing Interests**
The authors declare no competing interests.

**Data availability**
All data generated or analysed are given in the article.